\documentclass[preprint,5p,times,twocolumn]{elsarticle}
\usepackage{amssymb}
\usepackage{amsmath,amsfonts}
\usepackage{algorithmic}
\usepackage{algorithm}
\usepackage{array}
\usepackage{textcomp}
\usepackage{stfloats}
\usepackage{url}
\usepackage{verbatim}
\usepackage{graphicx}
\usepackage{amssymb}
\usepackage{amsmath}
\usepackage{multirow}
\usepackage{graphicx}
\usepackage{booktabs}
\usepackage{bbding}
\usepackage[colorlinks,
            linkcolor=green,
            anchorcolor=green,
            citecolor=green]{hyperref}

\journal{}

\begin{document}

\begin{frontmatter}

\title{Variable Aperture Bokeh Rendering via Customized Focal Plane Guidance}

\author[1]{Kang Chen\fnref{equA}} 
\author[1]{Shijun Yan\fnref{equA}}
\author[2,1]{Aiwen Jiang\corref{cor1}}
\ead{jiangaiwen@jxnu.edu.cn}

\author[1]{Han Li}
\author[3]{Zhifeng Wang}

\address[1]{School of Computer and Information Engineering, Jiangxi Normal University, Nanchang 330022, China}
\address[2]{School of Digital Industry, Jiangxi Normal University, Shangrao 333400, China}
\address[3]{School of Computer Science, National University of Defense Technology, Changsha 410073, China}

\cortext[cor1]{Corresponding author.}
\fntext[equA]{Both contributed equally to this work.}

\begin{abstract}
Bokeh rendering is one of the most popular techniques in photography. It can make photographs visually appealing, forcing users to focus their attentions on particular area of image. However, achieving satisfactory bokeh effect usually presents significant challenge, since mobile cameras with restricted optical systems are constrained, while expensive high-end DSLR lens with large aperture should be needed.  Therefore, many deep learning-based computational photography methods have been developed to mimic the bokeh effect in recent years. Nevertheless, most of these methods were limited to rendering bokeh effect in certain single aperture. There lacks user-friendly bokeh rendering method that can provide precise focal plane control and customised bokeh generation. There as well lacks authentic realistic bokeh dataset that can potentially promote bokeh learning on variable apertures. To address these two issues, in this paper, we have proposed an effective controllable bokeh rendering method, and contributed a Variable Aperture Bokeh Dataset (VABD). In the proposed method, user can customize focal plane to accurately locate concerned subjects and select target aperture information for bokeh rendering. Experimental results on public EBB! benchmark dataset and our constructed dataset VABD have demonstrated that the customized focal plane together aperture prompt can bootstrap model to simulate realistic bokeh effect. The proposed method has achieved competitive state-of-the-art performance with only 4.4M parameters, which is much lighter than mainstream computational bokeh models. The contributed dataset and source codes will be released on github \href{https://github.com/MoTong-AI-studio/VABM}{https://github.com/MoTong-AI-studio/VABM}.
\end{abstract}


\end{frontmatter}

\section{Introduction}
\label{sec:intro}
What photographers call Bokeh, is a photographic effect that refers to the aesthetic quality of out-of-focus areas in image~\cite{wright2016photography}. Bokeh becomes one of the most popular subjects in photography, because it can make photographs visually appealing, forcing us to focus our attention on a particular area of the image. 

Bokeh Rendering, also known as shallow depth of field processing, can achieve soft beauty and layering. However, as defined, Bokeh does not refer to the blur itself or the amount of blur in the foreground or the background. In most case, it refers to the quality and feel of the background/foreground blur and reflected points of light. In practice, it presents significant challenge to obtain bokeh effects, since expensive professional DSLR camera lenses with large aperture are needed. 

\begin{figure}[h]
  \includegraphics[width=\linewidth]{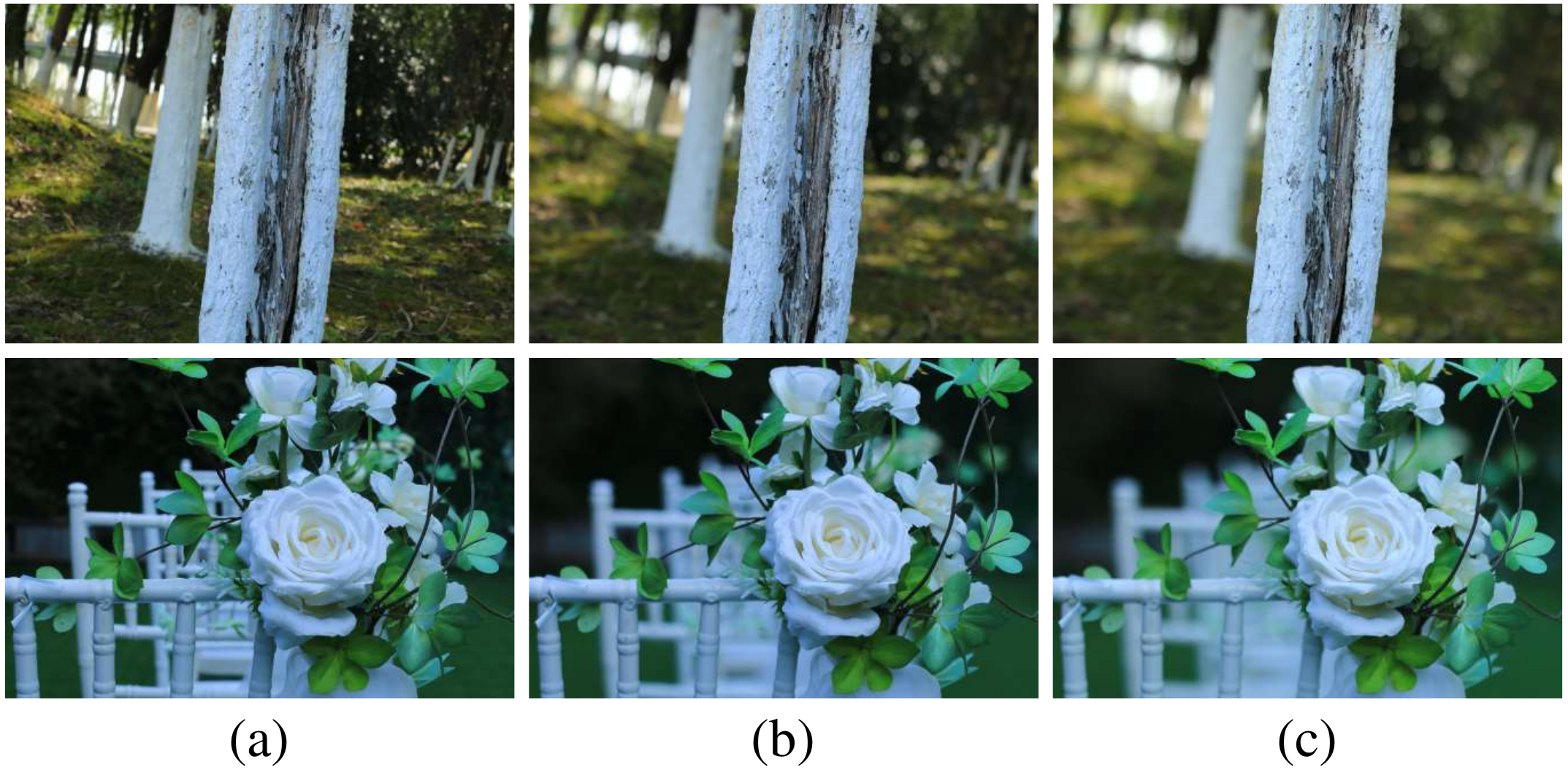}
  \caption{The effect of different aperture sizes on bokeh effects. (a) Narrow aperture "f/8". (b) Aperture "f/2.8". (c) Large aperture "f/1.8". The larger the aperture, the more pronounced bokeh effect.}
  \label{fig:A-B}
\end{figure}

Generally, portrait and telephoto lenses with large maximum apertures yield more pleasant-looking bokeh than cheaper consumer zoom lenses. As illustrated in Figure~\ref{fig:A-B}, smaller aperture serves to diminish bokeh effect, whereas larger aperture has obvious bokeh effect. Mobile cameras with restricted optical systems constrains generating natural bokeh effects. 

The contradiction between user's aesthetic requirements on photo quality and expensive high-end SLR cameras exists. Therefore, computational bokeh rendering has emerged as an attractive computer vision technology for engineering applications on imaging systems~\cite{conde2023perceptual,dutta2021depth,ignatov2022efficient,ignatov2020rendering}.

In recent years, many impressive bokeh rendering algorithms have been proposed. Principally, these methods can be classified into two categories: automatic bokeh rendering~\cite{ignatov2020rendering, qian2020bggan, dutta2021stacked, wang2022self, luo2023refusion} and controllable bokeh rendering~\cite{purohit2019depth, huang2023natural, yang2023bokehornot, seizinger2023efficient}. 

Specifically, automatic bokeh rendering strategy typically relies on end-to-end single model that automatically processes focusing and blurring without additional input information specific to real bokeh imaging. Though simple the network structure is, it fails to achieve perfect bokeh on background while keeping foreground subject intact.

The motivation of controllable bokeh rendering strategy is to provide a user-friendly interactive experience through which users can precisely control model's inputs and customise bokeh that meets specific visual needs and aesthetic requirements. To achieve this object, model requires additional input information to ensure accurate bokeh effect generation. Currently, parallax and image depth are popularly employed information. With these information, controllable strategy can yield more superior visual quality. Inevitably, it as well presents great challenges, such as high computational costs of depth estimation and difficulties on parallax information acquisition.

In spite of that, current mainstream controllable strategies regretfully still cannot meet the requirements of accurate control. For example, there lacks model that cherishes user-interaction for personalized focal plane selection. Without user-confirmed focal plane and depth-of-field, model cannot customise where needs bokeh. Moreover, there lack available dataset that helps promote research on controllable Bokeh in real-world scenes, especially, lacking appropriate dataset for multi-aperture bokeh switching. Even the EBB! benchmark bokeh dataset~\cite{ignatov2020rendering} that has already been widely accepted cannot meet this requirement. 

In order to address the aforementioned problems, we have proposed an efficient and interactive bokeh rendering method designated as Variable Aperture Bokeh Model (VABM). The proposed model receives depth maps, a preliminary user-provided mask of focal plane, and optional target aperture information to render bokeh. Specifically, the user-supplied mask, in combination with the estimated depth map, isolates the desired focal plane position, thereby more accurately determining the position of the subject, allowing the subject of the rendered image to remain unchanged. In order to reduce computational costs associated with the bokeh rendering process, Mamba~\cite{gu2023mamba} is employed to establish long-range dependencies within image content.

Furthermore, to promote multi-aperture bokeh research, we have carefully contributed a Variable Aperture Bokeh Dataset (VABD) which contains paired images with varying aperture sizes. The purpose of this dataset is to enable model to achieve controlled bokeh between different apertures. In contrast to synthetic datasets such as the Bokeh Effect Transformation Dataset (BETD)~\cite{conde2023lens}, all paired images in VABD dataset are derived from real-world data and conform to physical imaging principles, without artificially incorporating lens transition data. 

The proposed VABM network has been evaluated and compared with several state-of-the-art methods on publicly available large-scale bokeh dataset "EBB!"\cite{ignatov2020rendering}, as well as our contributed VABD dataset. The experimental results demonstrate that the proposed VABM can achieve superior results with much lightweight computation burdens.

The contributions of this paper are summarized as followings:
\begin{itemize}
\item We have proposed an effective and lightweight controllable bokeh rendering network. In the proposed network, users can interactively and conveniently ascertain the located focal plane, select target aperture information for bokeh rendering. User-guided focal plane together aperture prompt can bootstrap the model to simulate realistic bokeh effect that is consistent with physical imaging principles.
\item We have contributed a realistic multi-aperture bokeh rendering dataset. It is the first non-synthetic bokeh dataset with multi-aperture lens information, which fills the gap in the research field of realistic controllable bokeh switching.
\item We have pioneered the introduction of Mamba into bokeh rendering task, and constructed a strong baseline for our contributed dataset. Experiment results have demonstrated that the proposed model can achieve superior performance with lightweight computation burdens, compared with mainstream popular state-of-the-art models.
\end{itemize}

\section{Related work}
\subsection{Bokeh Rendering}
Computational bokeh rendering is an attractive computer vision technique for applications on imaging. We roughly categorize the current mainstream bokeh rendering methods into two strategies: automatic bokeh rendering and controllable bokeh rendering. 

\subsubsection{Automatic bokeh rendering}
Automatic rendering identifies all-in-focus foreground and out-of-focus background without user-interaction. Typically, Ignatov et al.~\cite{ignatov2020rendering} proposed a multi-scale end-to-end bokeh rendering network called PyNet by utilising image captured with narrow aperture and pre-computed depth-map. They has simultaneously released a large-scale bokeh dataset called EBB! (Everything is Better with Bokeh!). The dataset is consisting of 5000 image pairs with shallow and wide depth-of-field using Canon 70D digital SLR camera with prime lens (50mm f/1.8). Many following bokeh rendering work were evaluated on the "EBB!" dataset. 

Qian et al.~\cite{qian2020bggan} proposed a GAN-based approach to improve the visual quality of bokeh rendering and won the first place in the Rendering Realistic Bokeh Challenge track in 2020. Dutta et al.~\cite{dutta2021stacked} proposed a deep multi-scale hierarchical network without any monocular depth estimation module. It greatly reduced model's parameter count and runtime. Wang et al.~\cite{wang2022self} proposed an efficient multi-scale pyramid fusion network. Luo et al.~\cite{luo2023refusion} employed diffusion model for bokeh rendering in latent feature space, which drastically reduced computational cost and improved network stability.

\subsubsection{Controllable Bokeh Rendering}
Controllable bokeh rendering requires personalized information as extra inputs to customise bokeh effect. Typically, Purohit et al.~\cite{purohit2019depth} improved the aesthetic quality of bokeh rendering through introducing pre-trained salient region segmentation and depth estimation modules. Peng et al~\cite{Peng2022BokehMe} proposed a hybrid bokeh rendering framework called Bokehme. The main idea of Bokehme was to marry neural renderer with classical physically motivated renderer. Huang et al.~\cite{huang2023natural} proposed to take all-in-focus image and depth image as inputs to estimate circle-of-confusion parameters for each pixel. Yang et al.~\cite{yang2023bokehornot} proposed to embed lens metadata into model, guiding switching of bokeh effects between blur case and sharp case. Seizinger et al.~\cite{seizinger2023efficient} proposed an EBokehNet to embedding lens type and aperture information. 

In most case, depth-of-field information is required to realize controlled bokeh renderings~\cite{lee2021iterative, xu2021virtual, zhang2021pr, wan2017benchmarking}. The depth-of-field refers to the range within which user-concerned subjects being photographed are kept in focus. It is mainly affected by the aperture size of lens, the focal length and the shooting distance. Given certain lens and shooting distance, smaller aperture (larger f-number) produces wider depth-of-field, while larger aperture (smaller f-number) results shallower depth-of-field, thus exacerbating the Bokeh effect.

Automatic rendering strategy simplified rendering process at the cost of sacrificing reasonable explanation of physical imaging principles. As a result, methods of this category cannot accurately capture depth-of-field in real scenes, lacking of model flexibility and controllability. In this paper, besides convention depth-map and aperture embedding, we provide a convenient user-interaction way to realize personalized focal plane control, thereby producing better quality visual effects.

%\subsection{Depth Estimation}
%Depth maps are of great significance in computer vision applications. In this section, we briefly review some representative research that focused on estimating depth from single RGB image. Typically, Eigen et al.~\cite{eigen2014depth} employed two deep network stacks to address the uncertainty associated with depth estimation from a single image, involving global estimation of entire image and localized refinement. Fu et al~\cite{fu2018deep} introduced an increment discretization (SIG) strategy to achieve higher accuracy and faster synchronous convergence for single-image depth estimation.

%Choi et al.~\cite{choi2015depth} utilized a set of continuously captured RGB-D images in conjunction with gradient domain techniques to achieve high-quality bokeh rendering effects. Petland's work \cite{pentland1987new} on depth estimation using defocused images has been highly influential. Most of shape-fro-defocus (SFD) methods require one or more images captured from the same viewpoint and scene as the training data. Lin et al. \cite{lin2013absolute} proposed a method for estimating absolute depth from a single defocused image, building on the development of techniques for quantifying focus.

\subsection{State Space Models}
State Space Model (SSM)~\cite{hamilton1994state}, originally as mathematical framework to describe dynamic systems' behavior, in recent years, has been introduced into deep learning field~\cite{gu2021combining} for long-range modeling. SSM has achieved remarkable success in natural language processing and computer vision applications~\cite{gu2021efficiently, gupta2022diagonal, li2022makes, smith2022simplified}. 

%Representatively, Gu et al.~\cite{gu2021efficiently} were the pioneers who introduced Structured State Space Model (S4) in deep learning field. Building upon S4 model, Smith et al.~\cite{smith2022simplified} proposed S5 model which integrated Multi-Input Multi-Output (MIMO) SSM with efficient parallel scanning strategy, addressing the parallel computation limitations of S4.

Representatively, Gu et al.~\cite{gu2023mamba} have proposed in pioneer a selective state space model called Mamba. It has demonstrated that Mamba can achieve superior performance compared to existing Transformer-based methods across various natural language processing tasks~\cite{lieber2024jamba, patro2024simba, grazzi2024mamba}. Moreover, Mamba has been extensively utilized in computer vision tasks including image classification~\cite{liu2024vmamba, chen2024rsmamba}, medical image segmentation~\cite{liao2024lightm, sanjid2024integrating}, and image restoration~\cite{guo2024mambair, bai2024retinexmamba}. 

The application of Mamba has achieved great progress, especially on reduce model complexity. Therefore, in this paper, we incorporated it into the proposed bokeh rendering model as strong baseline.

\section{Method}
\subsection{Preliminaries}
\begin{figure}[h]
  \centering
  \includegraphics[width=\linewidth]{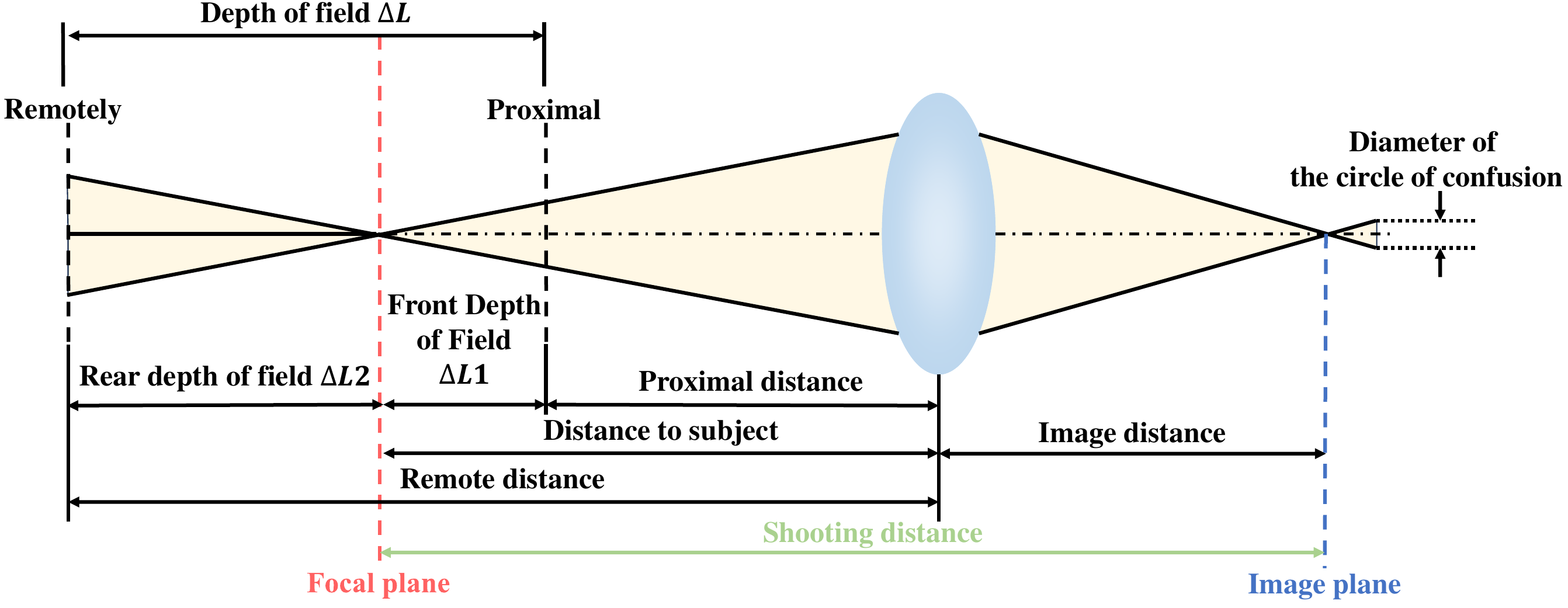}
  \caption{Illustration on physical imaging principles. Objects within certain depth ranges around focal plane can be clearly projected on image plane. Objects outside the focal plane will be out of focus, forming the observed confusion circle.}
  \label{fig:image_principle}
\end{figure}

In photography, subjects within depth-of-field are highlighted in focus. As illustrated in Figure~\ref{fig:image_principle}, when subject is located at focal plane, the area in front and behind it will appear blurred. These out-of-focus pixels spread on imaging plane to form circles of confusion (CoC). 

The size of CoC depends on aperture diameter, focal length and shooting distance. Mathematically, the formation of CoC is given in Equation~\ref{equ:coc}. The larger the aperture (i.e., the smaller the f-number), the shallower the depth of field, consequently the larger the CoC, and finally the more pronounced the bokeh effect. 
\begin{equation}
\label{equ:coc}
\begin{gathered}
r = \frac{A}{2} \times \frac{f}{D_o} \times \frac{|D_o - D_f|}{|D_f - f|}
\end{gathered}
\end{equation}
where, $r$ is the diameter of CoC, $f$ represents the focal length, $D_f$ is the focal plane distance, $A$ is aperture diameter, and $D_o$ is the object distance. 

Given a target lens with fixed focal length, if objective focal plane is confirmed, then the bokeh effects can be efficiently and realistically simulated by relative depth estimation and aperture diameter information. Foreground subjects can be better guaranteed to be focused. Therefore, it is crucial to determine appropriate focal plane for realistic bokeh effect simulation.

\subsection{Overall Pipeline}
\begin{figure*}[h]
  \centering
  \includegraphics[width=\linewidth]{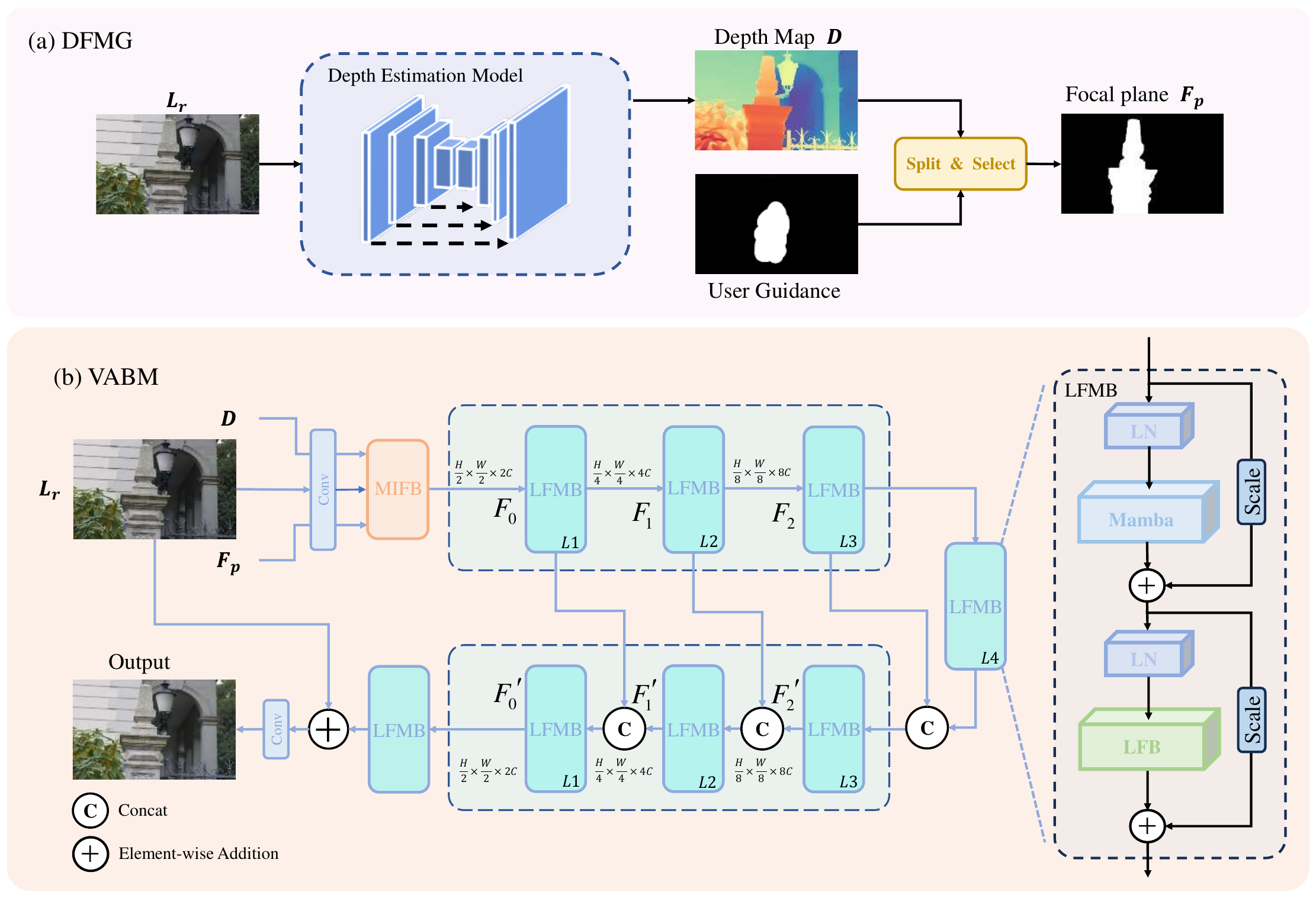}
  \caption{The framework of the proposed method. Depth map and focal plane are generated in DFMG, which are then fed into the Variable Aperture Bokeh Model (VABM) for fusion, providing the model with information such as depth relationship and focused subject to guide the bokeh rendering process. And the target lens information is fully fused with the features in each LFMB to achieve the effect of selectable target lens.}
  \label{fig:model}
\end{figure*}

We have proposed an effective variable aperture bokeh model (VABM) for computational bokeh rendering. The overall pipeline is shown in Figure~\ref{fig:model}. It comprises two cascade stages. In the first stage, pretrained Depth-Anything-V2~\cite{depth_anything_v2} network is employed to estimate image-level depth. Then focal plane is adaptively confirmed on the estimated depth map through a user-provided mask. In the second stage, an efficient multi-scale mamba-based lightweight model is proposed to fuse information including depth map, ascertained focal plane, and aperture to simulate realistic bokeh effect. 

The proposed network has realized bokeh rendering with customized target aperture. In the following subsections, we will describe the network structures in details.

\subsection{Focal Plane Customization}

In order to conveniently determine focal plane where the rendered subject is located, we have proposed a strategy for potential focal plane customization. Specifically, we compute the histogram of depth map $I_d$, and normalize it to obtain depth distribution $P_i$. Then we define the between-class variance for multiple thresholds and compute respective weights and means for each class. For every class $j$ between interval of threshold  $t_{j-1}$ and $t_j$, the weight $w_j$ and mean $\mu_j$ are calculated as following Equation~\ref{equ:weight_mean}:
\begin{equation}
\label{equ:weight_mean}
\begin{gathered}
w_j = \sum_{i=t_{j-1}+1}^{t_j} P(i), \\
\mu_j = \frac{\sum_{i=t_{j-1}+1}^{t_j} i \cdot P(i)}{w_j}
\end{gathered}
\end{equation}

To maximize between-class variance and find optimal threshold, we optimize the process as defined in Equation~\ref{equ:otsu}:
\begin{equation}
\label{equ:otsu}
(t_1^*, t_2^*, \ldots, t_{K-1}^*) = \arg\max_{t_1, t_2, \ldots, t_{K-1}} \sum_{j=0}^{K-1} w_j (\mu_j - \mu_T)^2.
\end{equation}
where, $\mu_T$ is global average depth, $(t_1^*, t_2^*, \ldots,t_{K-1}^*)$ is the optimal thresholds, $k$ is the number of classes.

Through applying these optimal thresholds on depth map $I_d$, $K$ segmentation regions are generated, each of which can be taken as a potential focal plane. Then, user-provided mask is mapped to one of the regions. In order to ensure that the depth range of user-concerned subject can be covered as much as possible, $K$ is empirically set to 3. Consequently, the selected region is approximated as focal plane where the subject is located. The selected depth range is approximated as corresponding depth-of-field. Guided by the focal plane prompt, the model can more accurately locate user-concerned subject and better render background blurrings.

\subsection{Multiple Information Fusion Block}
As we know, image-level depth map contains global depth distributions of all subjects in imaging system, while the selected focal plane is concentrated on local region where concerned foreground subjects are located. In order to make full use of global depth map as well as local focal plane, we propose a Multiple Information Fusion Block (MIFB), whose structure is as shown in Figure~\ref{fig:mifb&lfb}:
\begin{figure*}[h]
  \includegraphics[width=\linewidth]{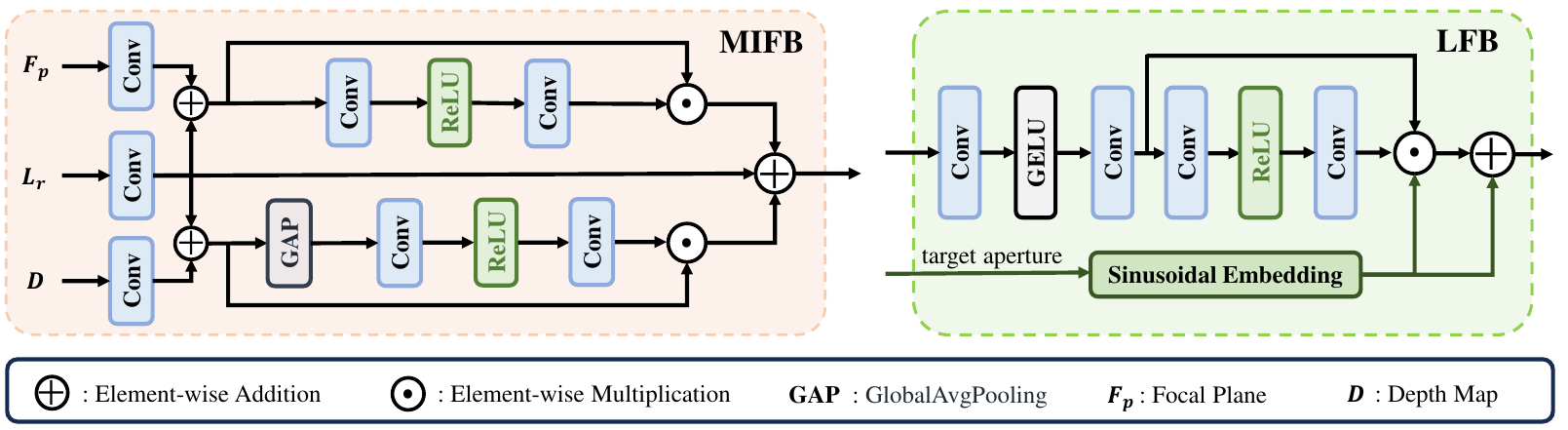}
  \caption{Structures of Multiple information fusion block (MIFB) and Lens Fusion Block (LFB). MIFB extracts and fully fuses the global information of the depth map and the local information of the focal plane to provide guidance for subsequent processes. LFB embeds the lens information into the model through sinusoidal embedding to distinguish different target lenses.}
  \label{fig:mifb&lfb}
\end{figure*}

\begin{figure*}[h]
  \includegraphics[width=\linewidth]{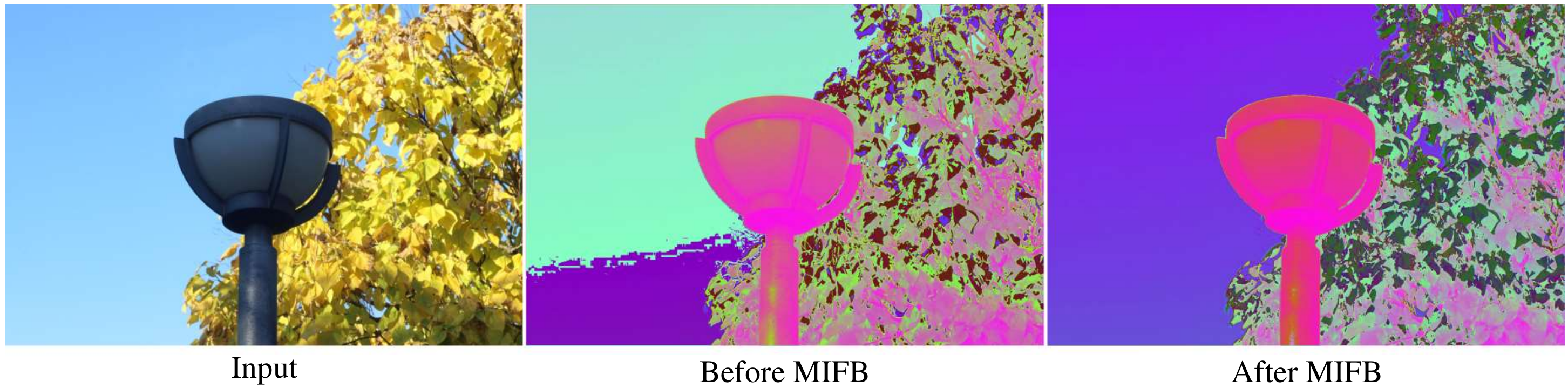}
  \caption{Example on the effectiveness of MIFB. the "Before MIFB" is the feature map obtained without fusing the depth of field map and focal plane, while the "After MIFB" is the feature map obtained after fusion.Compared with Before MIFB, the attention of the sky background with similar depth of field in After MIFB is closer, and more attention is paid to the location of the subject.}
  \label{fig:feature_constract}
\end{figure*}

Specifically, depth map $I_{d}$ and focal plane $F_p$ are integrated with all-in-focus image $L_r$ to obtain respective feature maps $x_g$ and $x_l$. Then, they are processed to derive global attention weights $w_g$ and local attention weights $w_l$. These weights respectively highlight corresponding depth distributions on all-in-focus feature map, obtaining accurate and user-concerned depth distributions. 

The information fusion process is formulated as in Equation~\ref{equ:5}:
\begin{equation}
\label{equ:5}
\begin{gathered}
x = Conv(L_r), x_g = x + Conv(I_{d}), x_l = x + Conv(F_p), \\
w_g = \sigma(BN(Conv(ReLU(BN(Conv(GAP(x_g))))))),\\
w_l = \sigma(BN(Conv(ReLU(BN(Conv(x_l))))))\\
L_{out} = x_g \odot w_g + x_l \odot w_l + x.
\end{gathered}
\end{equation}
where, BN indicates batch normalization. $\odot$ represents element-wise multiplication. GAP refers to global average pooling in channel direction. $\sigma$ represents sigmoid activation.

As illustrated in Figure~\ref{fig:feature_constract}, it can be observed that the fusion block enables model better distinguish different depth of field areas. The focal plane guidance further improves the accuracy of foreground recognition, so that it can better ensure that foreground subjects to be focused are not affected by background blurring.

\subsection{Lens-Fusion Mamba Block}
To empower our model has the ability to deal with high-resolution all-in-focus images, we have designed a Lens-Fusion Mamba Block (LFMB). The block structure is demonstrated in Figure~\ref{fig:model}. 

Specifically, Mamba is firstly employed to establish long-distance dependencies for input feature map $X_{i}$. Then target lens information are fused within Lens-Fusion sub-Block (LFB) through channel interaction. Benefit from the satisfied linear time complexity of Mamba, the proposed LFMB can efficiently avoid high computational costs and memory overhead.

The information processing pipeline of LFMB is defined in Equation~\ref{eq:lfmb}.
\begin{equation}
\label{eq:lfmb}
\begin{gathered}
    Z_i = Mamba(LN(X_i))+ s_1 \cdot X_{i}, \\
    X_o = LFB(LN(Z_i)) + s_2 \cdot Z_i
\end{gathered}
\end{equation}
where, $s_1$ and $s_2 \in \mathbb{R}^N$ are adjustable scaling factor, $LN$ represents layer normalization.

The detailed process of $LFB$ for lens fusion is formulated in Equation~\ref{equ:lfb}:
\begin{equation}
\label{equ:lfb}
\begin{gathered}
y = Conv(GELU(Conv(y_i))), \\
w = \sigma(Conv(ReLU(Conv(GAP(y))))),\\
y_{o} = (y \odot w) \odot lens + lens.
\end{gathered}
\end{equation}
where, $lens$ is sinusoidal embedding of target aperture. GAP is global average pooling in channel direction.

\begin{figure*}[!h]
  \centering
  \includegraphics[width=\linewidth]{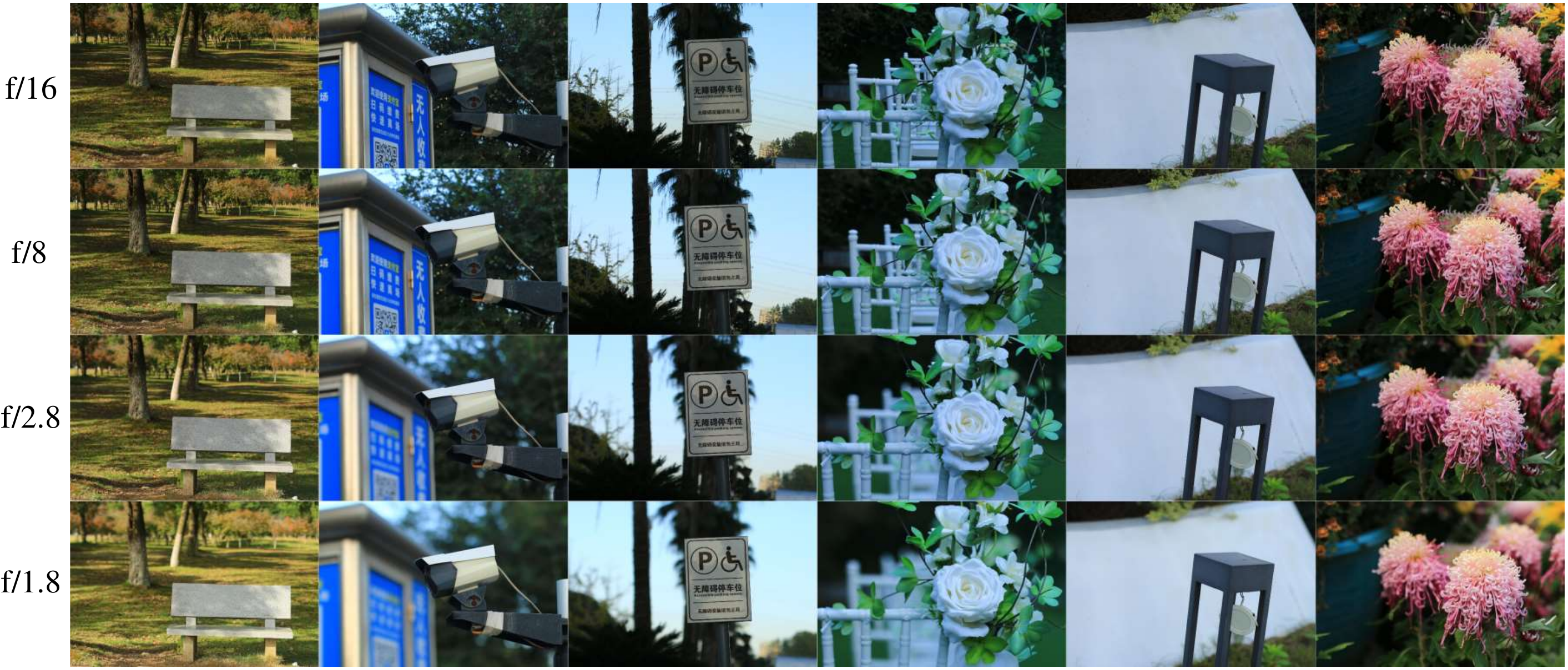}
  \caption{ A sample extracted from the VABD dataset showing, from left to right, six different scenes selected with aperture sizes of \(f/1.8\), \(f/2.8\), \(f/8\) and \(f/16\) for the same scene shot from top to bottom. Better observation if zoom in.}
  \label{fig:our_dataset}
\end{figure*}

\section{Experiments and Analysis}
\subsection{Dataset}
\subsubsection{EBB!}
The dataset EBB!(Everything is Better with Bokeh!)~\cite{ignatov2020rendering} was released on Bokeh Effects Rendering Challenge 2020 as benchmark dataset. Its training set contains 4600 images pairs, while its validation set and test set respectively contain 200 image pairs. 

Following common experiment settings, we uniformly adjusted image size of training data to $1536\times1024$ through Bilinear Interpolation. During training, we further randomly crop images to $1024\times1024$ patch size. Images in validation set and test set were retained in original size. 

%In order to ensure that the model performance is not affected, we deleted the obviously incomplete images in the EBB! dataset.

\subsubsection{VABD}
In order to promote research on computational bokeh rendering, We have contributed a new Bokeh dataset VABD (Variable Aperture Bokeh Dataset). It includes both indoor and outdoor high-resolution images captured by Canon lenses with different aperture size. Therefore images in VABD dataset were captured with different depth-of-field. 

Specifically, four different apertures (f/1.8, f/2.8, f/8.0 and f/16.0) are employed to capture images of the same scene by using the same camera lens. As a result, in VABD dataset, totally 535 image groups with the four different aperture cases were taken as training set. 200 image groups with three different apertures (f/1.8, f/2.8 and f/8.0) were taken as test set. All these images were taken at different locations, under different lighting conditions and weather conditions. In order to ensure content consistency of image pairs captured in the same scene and to avoid image quality degradation due to slight changes on camera viewpoint, tripod is utilized to stabilize camera during image acquisition process. All default settings except aperture size are fixed throughout the acquisition process.

Taking into account the subtle uncertainty of real-world image shooting practice, following EBB!~\cite{ignatov2020rendering}, we have performed the same SIFT and RANSAC techniques on paired images for image registration. Finally, after registration, images were uniformly cropped to a size of $1536\times1024$. Some representative sample images in VABD are presented in Figure~\ref{fig:our_dataset}.

To the best of our knowledge, VABD is the first realistic bokeh dataset that contains paired bokeh images of real-world scenes in multiple aperture conditions.

\subsection{Experimental Setup}

\subsubsection{Implementation details}
The VABM is implemented by using PyTorch and trained by using AdamW as optimizer. During training phase, we apply random-cropping on input all-in-focus image to generate training patches in resolution of 1024 × 1024. The batch size is set 2. The learning rate is set 2e-4. The training loss combines L1 loss and SSIM loss. All experiments are conducted on Ubuntu 20.04.4 system equipped with Intel Core i7-11700 CPU, 32GB RAM, and NVIDIA GeForce RTX 3090 GPU of 24GB RAM.

\begin{table*}[t]
\renewcommand\arraystretch{1.2}
\caption{\label{tab:EBB!} Quantitative comparison of the performance and parameters with mainstream SOTA models on EBB! dataset. The best and second-best results are \textbf{highlighted} and \underline{underlined}, respectively.}
\centering
\setlength{\tabcolsep}{8pt} 
\begin{tabular}{c|ccccc}
   \hline
   Methods & PSNR(↑) & SSIM(↑) & LPIPS(↓) & Par.(M)(↓) & FLOPs(G)(↓)\\
   \hline
   DDDF\cite{purohit2019depth} & 24.14 & 0.8713 & 0.2482 & 13.84 & 117.56\\
   PyNet\cite{ignatov2020rendering} & 24.21 & 0.8593 & 0.2219 & 47.5 & 15.0\\
   BGGAN\cite{qian2020bggan} & 24.39 & 0.8645 & 0.2467 & 10.84 & 24.9\\
   DMSHN-os\cite{dutta2021stacked} & 24.57 & 0.8558 & 0.2289 & \underline{5.42} & 42.82\\
   DMSHN\cite{dutta2021stacked} & 24.73 & 0.8619 & 0.2271 & 10.84 & 85.63\\
   MPFNet\cite{wang2022self} & 24.74 & 0.8806 & 0.2255 & 6.12 & 67.1\\
   CoCNet\cite{huang2023natural} & \underline{24.78} & 0.8604 & --- & 7.34 & ---\\
   BRViT\cite{nagasubramaniam2023bokeh} & 24.76 & \textbf{0.8904} & \textbf{0.1924} & 123.14 & 108.3\\
   VABM(Ours) & \textbf{24.83} & \underline{0.8815} & \underline{0.2169} & \textbf{4.4} & \textbf{9.9}\\
   \hline
\end{tabular}
\end{table*}
\subsubsection{Evaluation metrics}

To evaluate the fidelity between the predicted results and the ground truth, peak signal-to-noise ratio (PSNR)~\cite{huynh2008scope}, structural similarity index (SSIM)~\cite{wang2004image}, and learned perceptual image patch similarity (LPIPS)~\cite{zhang2018unreasonable} are utilized as evaluation metrics. The number of model parameters and the computational cost on images of resolution $256 \times 256$ are employed as indicators to evaluate model complexity.

\begin{figure*}[!h]
  \centering
  \includegraphics[width=\linewidth]{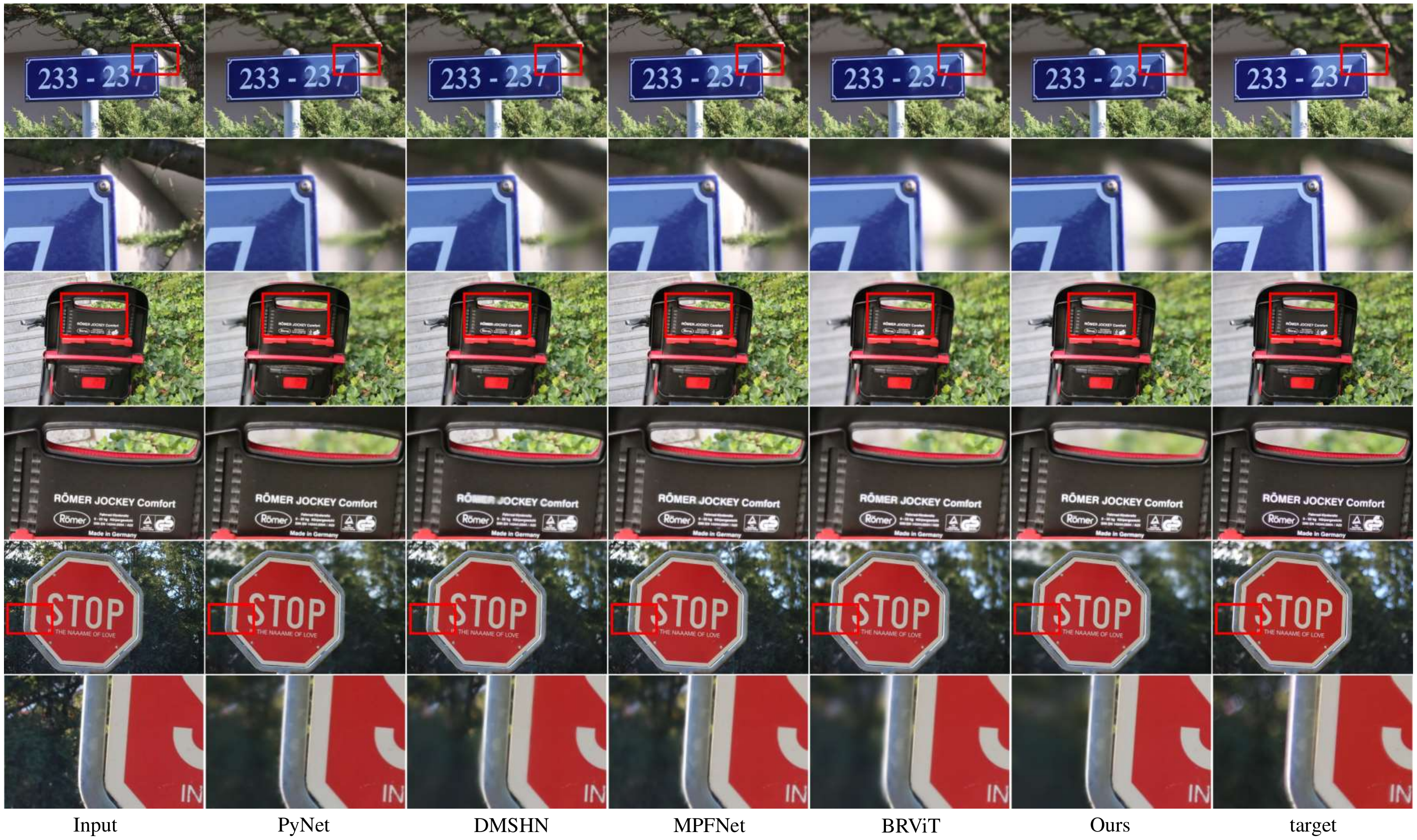}
  \caption{Visual comparisons on EBB! dataset. From left to right: PyNet\cite{ignatov2020rendering}, DMSHN-os, DMSHN\cite{dutta2021stacked}, MPFNet\cite{wang2022self}, and our VABM. Our method simulates a more realistic bokeh effect compared to the other methods, and can better preserve clear focused subjects. Better observation if zoom in.}
  \label{fig:ebb!}
\end{figure*}

\begin{table*}[!t]
\renewcommand\arraystretch{1.4}
\caption{\label{tab:VABD_one-by-one} \textbf{Under single-target apertures}, quantitative comparison of the performance and parameters with mainstream SOTA models on VABD dataset from all-in-focus image (of aperture f/16.0) to different bokeh effects (of respective apertures f/1.8, f/2.8, and f/8.0). The best and second-best results are \textbf{highlighted} and \underline{underlined}, respectively.}
\centering
\setlength{\tabcolsep}{5pt} 
\begin{tabular}{c|ccc|cc}
   \hline
   \multirow{2}{*}{Methods}  & Aperture=f/1.8 & Aperture=f/2.8 & Aperture=f/8 & \multirow{2}{*}{Par.(M)} & \multirow{2}{*}{FLOPs(G)}\\
    & PSNR/SSIM/LPIPS & PSNR/SSIM/LPIPS & PSNR/SSIM/LPIPS  \\
   \hline
   IR-SDE\cite{luo2023refusion} & 25.27/0.8543/0.3105 & 27.82/0.8667/0.2886 & 31.08/0.8805/0.2717 & 135.3 & 119.1\\
   DMSHN-os\cite{dutta2021stacked} & 25.59/0.8704/0.2943 & 27.91/0.8759/0.2839 & 31.19/0.8893/0.2711 & \underline{5.42} & \underline{42.82}\\
   DMSHN\cite{dutta2021stacked} & 25.92/0.8783/0.2886 & 28.03/0.8804/0.2791 & 31.31/0.8823/0.2707 & 10.84 & 85.63\\
   BokehOrNot\cite{yang2023bokehornot} & 25.97/0.8693/0.2903 & 28.15/0.8757/0.2796 & 31.47/\textbf{0.8866}/0.2653 & 34.4 & 130.4\\
   EBokehNet\cite{seizinger2023efficient} & 26.11/0.8802/\underline{0.2849} & 28.29/0.8814/0.2751 & \underline{31.54}/0.8839/0.2603  & 20.0 & 222.9\\
   BRViT\cite{nagasubramaniam2023bokeh} & \underline{26.17}/\underline{0.8811}/0.2853 & \underline{28.31}/\textbf{0.8827}/\underline{0.2742} & 31.53/0.8852/\textbf{0.2523} & 123.14 & 108.3\\
   VABM(Ours) & \textbf{26.23}/\textbf{0.8815}/\textbf{0.2844} & \textbf{28.37}/\underline{0.8823}/\textbf{0.2702} & \textbf{31.63}/\underline{0.8859}/\underline{0.2528} & \textbf{4.4} & \textbf{9.9}\\
   \hline
\end{tabular}
\end{table*}

\begin{figure*}[!h]
  \centering
  \includegraphics[width=\linewidth]{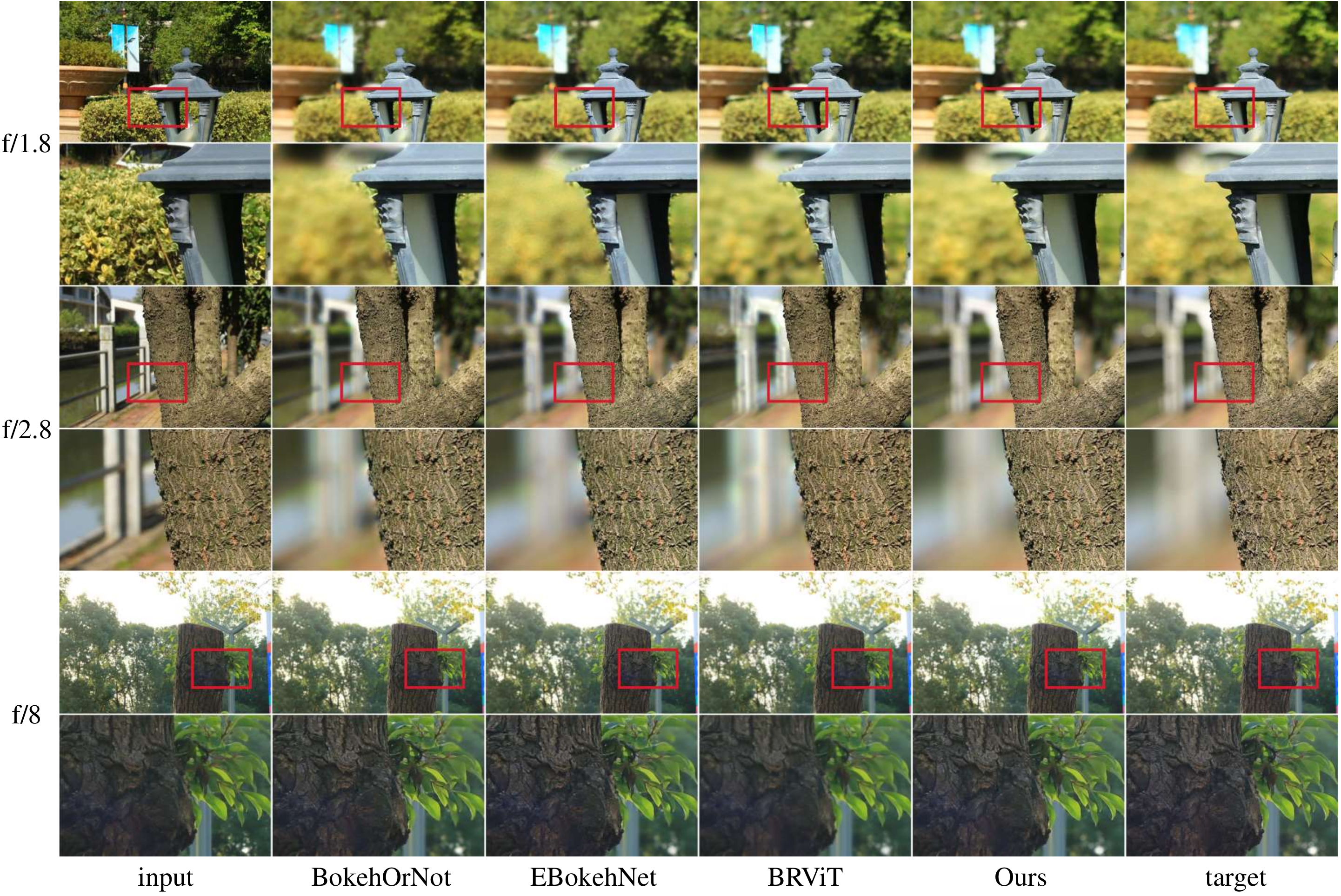}
  \caption{Visual comparisons on VABD dataset. From left to right, BokehOrNot\cite{yang2023bokehornot}, EBokehNet\cite{seizinger2023efficient}, BRViT\cite{nagasubramaniam2023bokeh} and our VABM. From top to bottom are images with aperture sizes of \(f/1.8\), \(f/2.8\), and \(f/8\). Better observation if zoom in.}
  \label{fig:VABD_one-by-one}
\end{figure*}

\begin{table*}[!t]
\renewcommand\arraystretch{1.4}
\caption{\label{tab:VABD_all-in-one} \textbf{Under multiple target apertures}, quantitative comparison of the performance and parameters with mainstream SOTA models on VABD dataset from all-in-focus image (of aperture f/16.0) to different bokeh effects (of respective apertures f/1.8, f/2.8, and f/8.0). The best and second-best results are \textbf{highlighted} and \underline{underlined}, respectively.}
\centering
\setlength{\tabcolsep}{5pt} 
\begin{tabular}{c|ccc|cc}
   \hline
   \multirow{2}{*}{Methods}  & Aperture=f/1.8 & Aperture=f/2.8 & Aperture=f/8 & \multirow{2}{*}{Par.(M)} & \multirow{2}{*}{FLOPs(G)}\\
    & PSNR/SSIM/LPIPS & PSNR/SSIM/LPIPS & PSNR/SSIM/LPIPS  \\
   \hline
   IR-SDE\cite{luo2023refusion} & 24.88/0.8492/0.3132 & 27.61/0.8610/0.2917 & 30.80/0.8794/0.2722 & 135.3 & \underline{119.1}\\
   BokehOrNot\cite{yang2023bokehornot} & 25.88/0.8672/0.2933 & 28.09/0.8730/0.2831 & \underline{31.39}/\textbf{0.8857}/0.2685 & 34.4 & 130.4\\
   EBokehNet\cite{seizinger2023efficient} & \underline{25.93}/\underline{0.8766}/\textbf{0.2865} & \underline{28.16}/\underline{0.8778}/\underline{0.2791} & 31.32/0.8823/\underline{0.2677} & \underline{20.0} & 222.9\\
   VABM(Ours) & \textbf{26.09}/\textbf{0.8797}/\underline{0.2871} & \textbf{28.24}/\textbf{0.8795}/\textbf{0.2727} & \textbf{31.52}/\underline{0.8834}/\textbf{0.2557} & \textbf{4.4} & \textbf{9.9}\\
   \hline
\end{tabular}
\end{table*}

\begin{figure*}[!h]
  \centering
  \includegraphics[width=\linewidth]{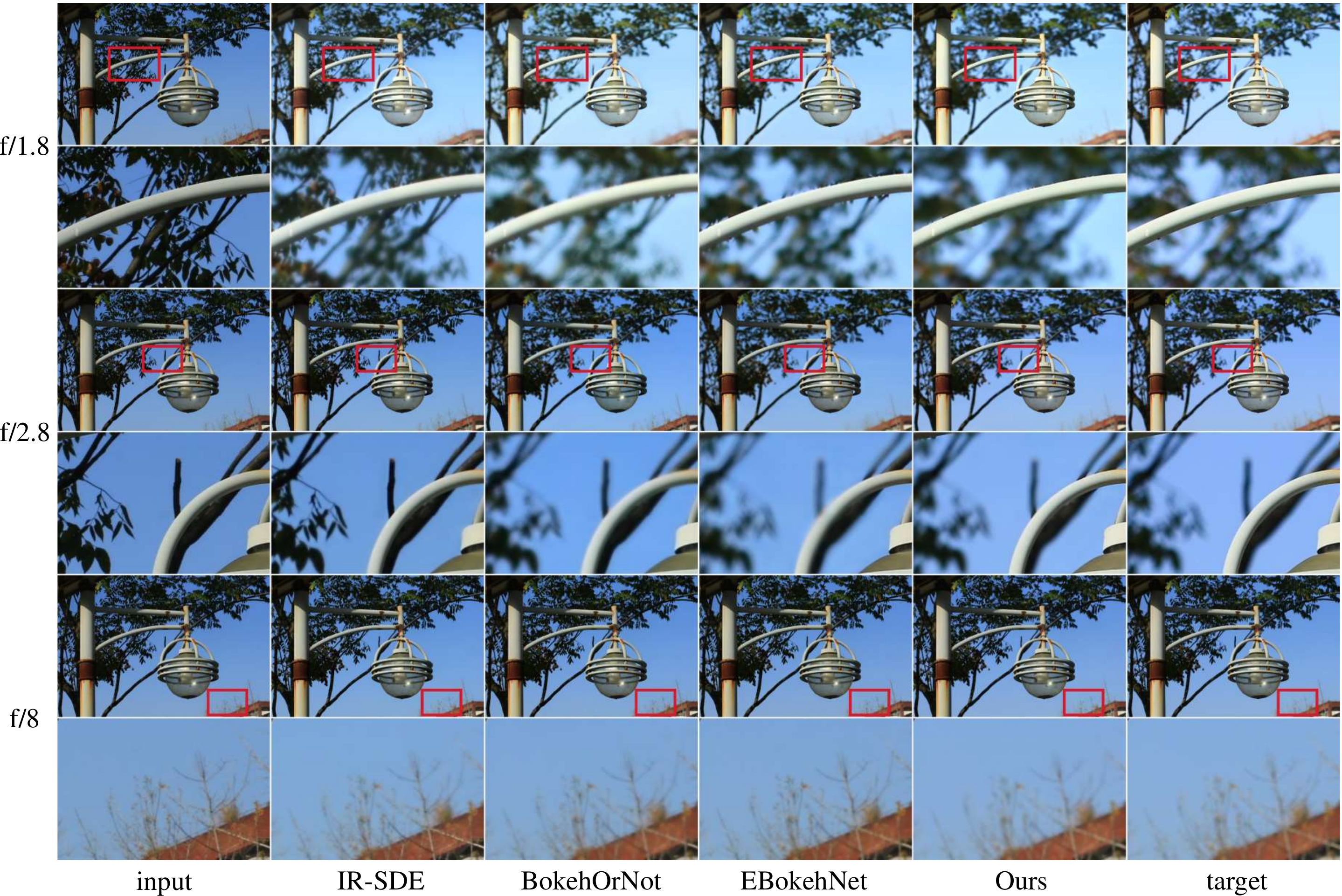}
  \caption{Visual comparison of previous classical models on the VABD dataset. From left to right, the original image with an aperture size of f/16, IR-SDE\cite{luo2023refusion}, BokehOrNot\cite{yang2023bokehornot}, EBokehNet \cite{seizinger2023efficient}, our VABM, and the target image . To facilitate the comparison between model performances, the three apertures are taken from the same scene and angle, and from top to bottom are images with aperture sizes of (f/1.8), (f/2.8), and (f/8), respectively. The results are better when viewed enlarged.}
  \label{fig:VABD_all-in-one}
\end{figure*}

% \subsection{Comparisons with State-of-the-art Methods}
\subsection{Comparison on the EBB! dataset}

We have compared VABM with eight representative methods including DDDF\cite{purohit2019depth}, PyNet\cite{ignatov2020rendering}, BGGAN\cite{qian2020bggan}, DMSHN\cite{dutta2021stacked}, DMSHN-os, MPFNet\cite{wang2022self}, CoCNet\cite{huang2023natural}, and BRVIT\cite{nagasubramaniam2023bokeh} on EBB! dataset. Due to the absence of pre-trained models or publicly available codes of some methods, for fair comparison, the scores provided by authors on EBB! test set were directly cited. 

The quantitative comparisons were presented in Table~\ref{tab:EBB!}. It can be observed that VABM outperformed majority of competing models on all metrics, except that VABM was slightly inferior to BRVIT on SSIM and LPIPS metrics. Notwithstanding, the parameter size of VABM was much smaller than that of BRVIT, approximately only 3.5\% of the parameter count of BRVIT model. The computational cost tested on $256 \times 256$ resolution images was only 9.9 GFLOPS, which was only 9.1\% of the computation cost of BRVIT. This fact indicated that our model was capable of maintaining high performance while having much fewer parameters than other models. 

To illustrate the efficacy of our model in bokeh rendering, we have presented visual comparisons on several representative samples, as shown in Figure~\ref{fig:ebb!}. It can be observed that VABM achieved bokeh effect more closely aligned with real case. The focused subjects were as well much clearer.

\subsection{Comparisons on the VABD dataset}
In this section, we will compare the performance of each model under VABD using a single-target aperture setting (one model weight corresponds to one aperture type) and a multi-target aperture setting (one model weight corresponds to multiple aperture types) to comprehensively evaluate the performance of each model.

\subsubsection{Under single target aperture setting}

To further verify the adaptability of the proposed model for bokeh rendering under single target aperture setting, we have evaluated VABM and the models in table \ref{tab:EBB!} on our contributed VABD dataset. We also compare with several state-of-the-art works for aperture transitions, such as IR-SDE\cite{luo2023refusion}, BokehOrNot\cite{yang2023bokehornot}, and EBokehNet\cite{seizinger2023efficient}. For fair comparison, they were reproduced on VABD in the same experiment settings as VABM. It should be necessary to declare that, since most of those compared models on EBB! dataset were not fully open-sourced, in this section, we had to only reproduce the open-sourced models for comparison on VABD.

The results of quantitative comparison on single-target aperture case were recorded in Table~\ref{tab:VABD_one-by-one}. More visual comparisons were shown in Figure~\ref{fig:VABD_one-by-one}. The comparison results have demonstrated that VABM achieved state-of-the-art performance with the fewest parameters and lowest computational burden. Compared to mainstream models, the proposed VABM has as well well handled the sharp-edge preservation on focused foregrounds and the bokeh effect on blurred backgrounds.

\subsubsection{Under multiple target apertures setting}

In the multi-aperture setting, we compared the few models capable of adapting to various aperture transitions, including IR-SDE \cite{luo2023refusion}, BokehOrNot \cite{yang2023bokehornot}, and EBokehNet \cite{seizinger2023efficient}.

The results of quantitative comparison on multi-target aperture case were recorded in Table~\ref{tab:VABD_all-in-one}. From the experiment results, we can observe that our model obtained the highest PSNR and SSIM values with smallest model size. It should be pointed that the parameters number and computation burden of our model were respectively only 22\% and 4.4\% of the second-best model EBokehNet. 

More visual comparisons were shown in Figure~\ref{fig:VABD_all-in-one}. The VABM conveniently integrated lens information, and achieved controllable bokeh rendering for different apertures, as well as better bokeh rendering effects.

\subsection{Ablation Studies}

\begin{table*}[!t]
\renewcommand\arraystretch{1.4}
\caption{\label{tab:ablation}
The ablation study for EBB! dataset and all-in-one settings of the VABD dataset. This is to investigate the necessity of each part in the network. The best and second-best results are \textbf{highlighted} and \underline{underlined}, respectively.}
\centering
\setlength{\tabcolsep}{5pt}
\begin{tabular}{c|ccc|ccc}
   \hline
   datasets & \multicolumn{3}{c|}{\textbf{VABD}} & \multicolumn{3}{c}{\textbf{EBB!}} \\
   \hline
   \multirow{2}{*}{Settings}  & Aperture=f/1.8 & Aperture=f/2.8 & Aperture=f/8 & \multirow{2}{*}{PSNR} & \multirow{2}{*}{SSIM} & \multirow{2}{*}{LPIPS}\\
    & PSNR/SSIM/LPIPS & PSNR/SSIM/LPIPS & PSNR/SSIM/LPIPS  \\
   \hline
   w/o MIFB & 25.53/0.8729/0.2946 & 27.67/0.8731/0.2797 & 30.83/0.8785/0.2703 & 24.39 & 0.8776 & 0.2322 \\
   w/o depth & 25.78/0.8776/0.2904 & 27.96/0.8777/0.2774 & 31.14/0.8808/0.2593 & 24.57 & 0.8792 & 0.2267 \\
   w/o focal & \underline{25.86}/\underline{0.8782}/\underline{0.2894} & \underline{28.04}/\underline{0.8779}/\underline{0.2743} & \underline{31.25}/\underline{0.8817}/\underline{0.2577} & \underline{24.61} & \underline{0.8793} & \underline{0.2259} \\
   VABM & \textbf{26.09}/\textbf{0.8797}/\textbf{0.2871} & \textbf{28.24}/\textbf{0.8795}/\textbf{0.2727} & \textbf{31.52}/\textbf{0.8834}/\textbf{0.2557} & \textbf{24.83} & \textbf{0.8815} & \textbf{0.2169} \\
   \hline
\end{tabular}
\end{table*}

To demonstrate the effectiveness of the respectively proposed modules in VABM, we have conducted ablation experiments on EBB! dataset and all-in-one settings on the VABD dataset. The ablation scenarios were set as followings: (1) "w/o MIFB": replacing MIFB module with direct addition for all information; (2) "w/o focal": without using focal plane; and (3) "w/o depth": without using depth map. 

As shown in Table~\ref{tab:ablation}, In case (1), without the MIFB module, the model’s performance on EBB! and VABM dropped significantly. The study indicated that fusing depth map and focal plane information was effective. Both had great impact on rendering. In cases (2) and (3), the model’s performance also dropped on average, which once again proved that the guidance of depth map and the focal plane enhanced model’s ability to learn realistic bokeh rendering.

%\section{Limitation and Future Work}
%Although VABM has achieved promising results across multiple datasets by leveraging Mamba's linear computational complexity and incorporating depth-of-field and focal plane information to guide bokeh rendering, there are still several aspects for improvement. For instance, further exploration of continuously differentiable lens transition process, and the adoption of more user-friendly interactive methods for focal plane separation could benefit enhancing model performance and practice applications. Additionally, we plan to expand the Variable Aperture Bokeh Dataset (VABD) in the future, providing a more robust data foundation for future bokeh rendering research.

\section{Conclusion}
In this paper, we have proposed a lightweight but effective controllable bokeh rendering network that can conveniently determine user-concerned focal plane with certain depth-of-field. We have moreover contributed a realistic Variable Aperture Bokeh Dataset (VABD) for exploring bokeh rendering with across-aperture transition. The experiments on available public EBB! dataset and our contributed VABD dataset, have demonstrated that the proposed model achieves state-of-the-art performance with much lower parameter size, outperforming mainstream state-of-the-art methods in terms of both quality and computational efficiency. Our proposed method and contributed dataset can serve as strong baseline for future research.

\bibliographystyle{elsarticle-num} 
\bibliography{ref}  
\end{document}